%% file: StoppingCriteria.tex
\documentclass{article}
\include{preamble}

\newcommand \E {\mathop{\mbox{\bf{E}}}\nolimits}

\renewcommand \Pr {\mathop{\mbox{\bf{P}}}\nolimits}
\newcommand \given {\mathrel{|}}
\newcommand \bz {\mathbf{z}}

\newcommand{\refalg}[1]{Alg.~\ref{#1}}
\newcommand{\reffig}[1]{Fig.~\ref{#1}}

\newcommand{\reffigseptwo}[2]{Fig.~\ref{#1},~\ref{#2}}

\newcommand{\refsec}[1]{Sec.~\ref{#1}}

\newcommand \wdbc {{\texttt{wdbc}}}
\newcommand \spam {{\texttt{spam}}}

\title{Cost-minimising strategies for data labelling : optimal
  stopping and active learning}

\author{Christos Dimitrakakis\\\texttt{christos.dimitrakakis@gmail.com}\\
  \and Christian Savu-Krohn\\\texttt{christian.savu-krohn@unileoben.ac.at}\\
  \\
  Chair of Information Technology, University of Leoben\\
  Leoben A-8700, Austria}

\begin{document}
%% \makeanontitle
\maketitle
\begin{abstract}
  Supervised learning deals with the inference of a distribution over
  an output or label space $\CY$ conditioned on points in an
  observation space $\CX$, given a training dataset $D$ of pairs in
  $\CX \times \CY$.  However, in a lot of applications of interest,
  acquisition of large amounts of observations is easy, while the
  process of generating labels is time-consuming or costly.  One way
  to deal with this problem is {\em active} learning, where points to
  be labelled are selected with the aim of creating a model with
  better performance than that of an model trained on an equal number
  of randomly sampled points.  In this paper, we instead propose to
  deal with the labelling cost directly: The learning goal is defined
  as the minimisation of a cost which is a function of the expected
  model performance and the total cost of the labels used.  This
  allows the development of general strategies and specific algorithms
  for
\begin{inparaenum}[(a)]
\item optimal stopping, where the expected cost dictates whether label
  acquisition should continue
\item empirical evaluation, where the cost is used as a performance
  metric for a given combination of inference, stopping and sampling methods.
\end{inparaenum}
Though the main focus of the paper is optimal stopping, we also aim to
provide the background for further developments and discussion in the
related field of active learning.
\end{abstract}

\section{Introduction}

Much of classical machine learning deals with the case where we wish
to learn a target concept in the form of a function $f : \CX \to \CY$,
when all we have is a finite set of examples $D=\{(x_i,
y_i)\}_{i=1}^n$.  However, in many practical settings, it turns out
that for each example $i$ in the set only the observations $x_i$ are
available, while the availability of observations $y_i$ is restricted
in the sense that either
\begin{inparaenum}[(a)]
\item they are only observable for a subset of the examples
\item further observations may only be acquired at a cost.
\end{inparaenum}
In this paper we deal with the second case, where we can actually
obtain labels for any $i \in D$, but doing so incurs a cost.  Active
learning algorithms (i.e.  \cite{cohn95active,icml:Schohn:2000}) deal
indirectly with this by selecting examples which are expected to
increase accuracy the most.  However, the basic question of whether
new examples should be queried at all is seldom addressed.

This paper deals with the labelling cost explicitly.  We introduce a
cost function that represents the trade-off between final performance
(in terms of generalisation error) and querying costs (in terms of the
number of labels queried).  This is used in two ways.  Firstly, as the
basis for creating cost-dependent stopping rules.  Secondly, as the
basis of a comparison metric for learning algorithms and associated
stopping algorithms.

To expound further, we decide when to stop by estimating the expected
performance gain from querying additional examples and comparing it
with the cost of acquiring more labels.  One of the main contributions
is the development of methods for achieving this in a Bayesian
framework.  While due to the nature of the problem there is potential
for misspecification, we nevertheless show experimentally that the
stopping times we obtain are close to the optimal stopping times.

We also use the trade-off in order to address the lack of a principled
method for comparing different active learning algorithms under
conditions similar to real-world usage.  For such a comparison a
method for choosing stopping times independently of the test set is
needed.  Combining stopping rules with active learning algorithms
allows us to objectively compare active learning algorithms for a
range of different labelling costs.

The paper is organised as follows.  Section~\ref{sec:cost} introduces
the proposed cost function for when labels are costly, while
Section~\ref{sec:metrics} discusses related work.
Section~\ref{sec:optimal_stopping} derives a Bayesian stopping method
that utilises the proposed cost function. Some experimental results
illustrating the proposed evaluation methodology and demonstrating the
use of the introduced stopping method are presented in
Section~\ref{sec:experiments}.  The proposed methods are not flawless,
however.  For example, the algorithm-independent stopping rule
requires the use of i.i.d.  examples, which may interfere with its
coupling to an active learning algorithm. We conclude with a
discussion on the applicability, merits and deficiencies of the
proposed approach to optimal stopping and of principled testing for
active learning.

\subsection{Combining Classification Error and Labelling Cost}
\label{sec:cost}
There are many applications where raw data is plentiful, but labelling
is time consuming or expensive.  Classic examples are speech and image
recognition, where it is easy to acquire hours of recordings, but for
which transcription and labelling are laborious and costly.  For this
reason, we are interested in querying labels from a given dataset such
that we find the optimal balance between the cost of labelling and the
classification error of the hypothesis inferred from the labelled
examples.  This arises naturally from the following cost function.

Let some algorithm $F$ which queries labels for data from some
unlabelled dataset $D$, incurring a cost $\gamma \in [0, \infty)$ for
each query.  If the algorithm stops after querying labels of examples
$d_1, d_2, \dotsc, d_t$, with $d_i \in [1, |D|]$.it will suffer a
total cost of $\gamma t$, plus a cost depending on the generalisation
error.  Let $f(t)$ be the hypothesis obtained after having observed
$t$ examples and corresponding to the generalisation error
$\E[R|f(t)]$ be the generalisation error of the hypothesis.  Then, we
define the total cost for this specific hypothesis as
\begin{equation}
  \label{eq:hypothesis_cost}
  \E[C_\gamma |f(t)] = \E[R|f(t)] + \gamma t.
\end{equation}
We may use this cost as a way to compare learning and stopping
algorithms, by calculating the expectation of $C_\gamma$ conditioned
on different algorithm combinations, rather than on a specific
hypothesis.

In addition, this cost function can serve as a formal framework for
active learning.  Given a particular dataset $D$, the optimal subset
of examples to be used for training will be $D^* = \argmin_i \E(R|F,
D_i) + \gamma |D_i|$.  The ideal, but unrealisable, active learner in
this framework would just use labels of the subset $D^*$ for training.

Thus, these notions of optimality can in principle be used both for
deriving stopping and sampling algorithms and for comparing them.
Suitable metrics of expected real-world performance will be discussed
in the next section.  Stopping methods will be described in
Section~\ref{sec:optimal_stopping}.

\subsection{Related Work}
\label{sec:metrics}
In the active learning literature, the notion of an objective function
for trading off classification error and labelling cost has not yet
been adopted.  However, a number of both qualitative and quantitative
metrics were proposed in order to compare active learning algorithms.
Some of the latter are defined as summary statistics over some subset
$\CT$ of the possible stopping times.  This is problematic as it could
easily be the case that there exists $\CT_1, \CT_2$ with $\CT_1
\subset \CT_2$, such that when comparing algorithms over $\CT_1$ we
get a different result than when we are comparing them over a larger
set $\CT_2$.  Thus, such measures are not easy to interpret since the
choice of $\CT$ remains essentially arbitrary.  Two examples are
\begin{inparaenum}[(a)]
\item the {\em percentage reduction in error}, where the percentage
  reduction in error of one algorithm over another is averaged over
  the whole learning curve~\cite{ijcai:saar:2001,icml:Melville:2004}
  and
\item the average number of times one algorithm is significantly
  better than the other during an arbitrary initial number of queries,
  which was used in \cite{ecml:Korner:2006}.
\end{inparaenum}
Another metric is the {\em data utilisation ratio} used in
\cite{ecml:Korner:2006, icml:Melville:2004, icml:abe:1998}, which is
the amount of data required to reach some specific error rate. Note
that the selection of the appropriate error rate is essentially
arbitrary; in both cases the concept of the {\em target error rate} is
utilised, which is the average test error when almost all the training
set has been used.  

Our setting is more straightforward, since we can use
\eqref{eq:hypothesis_cost} as the basis for a performance measure.
Note that we are not strictly interested in comparing hypotheses $f$,
but algorithms $F$.  In particular, we can calculate the expected cost
given a learning algorithm $F$ and an associated stopping algorithm
$Q_F(\gamma)$, which is used to select the {\em stopping time} $T$.
From this follows that the expected cost of $F$ when coupled with
$Q_F(\gamma)$ is
\begin{equation}
  \label{eq:expected_cost}
  v_e(\gamma, F, Q_F) \equiv  \E[C_\gamma |F, Q_F(\gamma)] = \sum_{t} \left(\E[R|f(t)] + \gamma t\right) \Pr[T=t \given F,Q_F(\gamma)]
\end{equation}

By keeping one of the algorithms fixed, we can vary the other in order
to obtain objective estimates of their performance difference.  In
addition, we may want to calculate the expected performance of
algorithms for a range of values of $\gamma$, rather than a single
value, in a manner similar to what~\cite{Bengio:Expected:2005}
proposed as an alternative to ROC curves.  This will require a
stopping method $Q_F(\gamma)$ which will ideally stop querying at a
point that minimises $\E(C_\gamma)$.

The stopping problem is not usually mentioned in the active learning
literature and there are only a few cases where it is explicitly
considered.  One such case is \cite{icml:Schohn:2000}, where it is
suggested to stop querying when no example lies within the SVM margin.
The method is used indirectly in \cite{icml:Campbell:2000}, where if
this event occurs the algorithm tests the current
hypothesis\footnote{i.e. a classifier for a classification task},
queries labels for a new set of unlabelled examples\footnote{Though
  this is not really an i.i.d.  sample from the original distribution
  except when $|D|-t$ is large.}  and finally stops if the error
measured there is below a given threshold; similarly,
\cite{icml:Balcan:2006} introduced a bounds-based stopping criterion
that relies on an allowed error rate.  These are reasonable methods,
but there exists no formal way of incorporating the cost function
considered here within them.  For our purpose we need to calculate the
expected reduction in classification error when querying new examples
and compare it with the labelling cost.  This fits nicely within the
statistical framework of optimal stopping problems.

\section{Stopping Algorithms}
\label{sec:optimal_stopping}
An optimal stopping problem under uncertainty is generally formulated
as follows.  At each point in time $t$, the experimenter needs to make
a decision $a \in A$, for which there is a {\em loss function}
$\CL(a|w)$ defined for all $w \in \Omega$, where $\Omega$ is the set
of all possible universes.  The experimenter's uncertainty about which
$w \in \Omega$ is true is expressed via the distribution
$\Pr(w|\xi_t)$, where $\xi_t$ represents his belief at time $t$.  The
{\em Bayes risk} of taking an action at time $t$ can then be written as
$\rho_0(\xi_t) = \min_a \sum_w \CL(a, w) \Pr(w|\xi_t)$.  Now, consider
that instead of making an immediate decision, he has the opportunity
to take $k$ more observations $D_k$ from a sample space $S^k$, at a
cost of $\gamma$ per observation, thus allowing him to update his
belief to $\Pr(w|\xi_{t+k}) \equiv \Pr(w|D_k, \xi_t)$.  What the
experimenter must do in order to choose between immediately making a
decision $a$ and continuing sampling, is to compare the risk of making
a decision now with the cost of making $k$ observations plus the risk
of making a decision after $k$ timesteps, when the extra data would
enable a more informed choice.  In other words, one should stop and
make an immediate decision if the following holds for all $k$:
\begin{equation}
  \label{eq:optimal_stopping}
  \rho_0(\xi_t) \leq \gamma k+
  \int_{S^k} p(D_k\!=\!s | \xi_t) \min_a \left[\sum_w \CL(a,w) \Pr(w | D_k\!=\!s, \xi_t)\right] \,ds.
\end{equation}
% {\tt[CD: or maybe this simpler formula?]}
% \[
% \min_a \E[R_t \given a, \xi_t] \leq \int_{S^k} p(D_k = s \given \xi_t) \min_a \E[R_{t+k} \given a, D_k=s, \xi_t] \,ds
% \]
We can use the same formalism in our setting. In one respect, the
problem is simpler, as the only decision to be made is when to stop
and then we just use the currently obtained hypothesis.  The
difficulty lies in estimating the expected error.  Unfortunately, the
metrics used in active learning methods for selecting new examples
(see \cite{ecml:Korner:2006} for a review) do not generally include
calculations of the expected performance gain due to querying
additional examples.

There are two possibilities for estimating this performance gain.  The
first is an algorithm-independent method, described in detail in
\refsec{sec:bayesian}, which uses a set of convergence curves, arising
from theoretical convergence properties.  We employ a Bayesian
framework to infer the probability of each convergence curve through
observations of the error on the next randomly chosen example to be
labelled.  The second method, outlined in
\refsec{sec:integrated_model}, relies upon a classifier with a
probabilistic expression of its uncertainty about the class of
unlabelled examples, but is much more computationally expensive.

\subsection{When no Model is Perfect: Bayesian Model Selection}
\label{sec:bayesian}
The presented Bayesian formalism for optimal sequential decisions
follows \cite{Degroot:OptimalStatisticalDecisions}.  We require
maintaining a belief $\xi_t$ in the form of a probability distribution
over the set of possible universes $w \in \Omega$.  Furthermore, we
require the existence of a well-defined cost for each $w$.  Then we
can write the Bayes risk as in the left side of
\eqref{eq:optimal_stopping}, but ignoring the minimisation over $A$
as there is only one possible decision to be made after stopping,
\begin{equation}
  \label{eq:bayes_risk}
  \rho_0(\xi_t) = \E(R_t \given \xi_t) = \sum_{w \in \Omega} \E(R_t \given w) \Pr(w \given \xi_t),
\end{equation}
which can be extended to continuous measures without difficulty.  We
will write the expected risk according to our belief at time $t$ for
the optimal procedure taking at most $k$ more samples as
\begin{equation}
  \label{eq:future_risk}
  \rho_{k+1}(\xi_t)  = \min \left\{ \rho_0(\xi_t), \E[\rho_k(\xi_{t+1}) \given \xi_t ] + \gamma \right\}.
  %% \quad \sum_{w \in \Omega} \E(R_{t+k} \given w) \Pr(w \given \xi_{t}) + \gamma k \right\}.
\end{equation}
This implies that at any point in time $t$, we should ignore the cost
for the $t$ samples we have paid for and are only interested in
whether we should take additional samples. The general form of the
stopping algorithm is defined in ~\refalg{al:bounded-stopping}.
\begin{algorithm}
  Given a dataset $D$ and any learning algorithm $F$, an initial
  belief $\Pr(w \given \xi_0)$ and a method for updating it,
  and additionally a known query cost $\gamma$, and a horizon $K$,
  \begin{algorithmic}[1]
    \FOR{$t=1, 2, \dotsc$}
    \STATE Use $F$ to query a new example $i \in D$ and obtain $f(t)$.
    \STATE Observe the empirical error estimate $v_t$ for $f(t)$.
    \STATE Calculate $\Pr(w \given \xi_t) = \Pr(w \given v_t, \xi_{t-1})$
    \IF {$\nexists$ $k \in [1, K]$ : $\rho_k(\xi_t) < \rho_0(\xi_t)$}
    \STATE Exit.
    \ENDIF
    \ENDFOR
  \end{algorithmic}
  \caption{A general bounded stopping algorithm using Bayesian inference.}
  \label{al:bounded-stopping}
\end{algorithm}
Note that the horizon $K$ is a necessary restriction for
computability.  A larger value of $K$ leads to potentially better
decisions, as when $K \to \infty$, the bounded horizon optimal
decision approaches that of the optimal decision in the unbounded
horizon setting, as shown for example in Chapter 12 of
\cite{Degroot:OptimalStatisticalDecisions}.  Even with finite $K>1$,
however, the computational complexity is considerable, since we will
have to additionally keep track of how our future beliefs $\Pr(w
\given \xi_{t+k})$ will evolve for all $k \leq K$.

\subsection{The {OBSV} Algorithm}
In this paper we consider a specific one-step bounded stopping
algorithm that uses independent validation examples for observing the
empirical error estimate $r_t$, which we dub OBSV and is shown in
detail in \refalg{al:obsv}.  The algorithm considers hypotheses $w \in
\Omega$ which model how the generalisation error $r_t$ of the learning
algorithm changes with time.  We assume that the initial error is
$r_0$ and that the algorithm always converges to some unknown
$r_\infty \equiv \lim_{t\to\infty} r_t$.  Furthermore, we need some
observations $v_t$ that will allow us to update our beliefs over
$\Omega$.  The remainder of this section discusses the algorithm in
more detail.

\subsubsection{Steps 1-5, 11-12.  Initialisation and Observations}
We begin by splitting the training set $D$ in two parts: $D_A$, which
will be sampled without replacement by the {\em active learning
  algorithm} (if there is one) and $D_R$, which will be {\em
  uniformly} sampled without replacement.  This condition is necessary
in order to obtain i.i.d. samples for the inference procedure outlined
in the next section.  However, if we only sample randomly, and we are
not using an active learning algorithm then we do not need to split
the data and we can set $D_A = \emptyset$.

At each timestep $t$, we will use a sample from $D_R$ to update
$p(w)$.  If we then expect to reduce our future error sufficiently, we
will query an example from $D_A$ using $F$ and subsequently update the
classifier $f$ with both examples.  Thus, not only are the
observations used for inference independent and identically
distributed, but we are also able to use them to update the classifier
$f$.

\subsubsection{Step 6.  Updating the Belief}
We model the learning algorithm as a process which asymptotically
converges from an initial error $r_0$ to the unknown final error
$r_\infty$.  Each model $w$ will be a {\em convergence estimate}, a
model of how the error converges from the initial to the final error
rate.  More precisely, each $w$ corresponds to a function $h_w :
\Naturals \to [0,1]$ that models how close we are to convergence at
time $t$.  The predicted error at time $t$ according to $w$, given the
initial error $r_0$ and the final error $r_\infty$, will be
\begin{equation}
  g_w(t \given r_0, r_\infty) = r_0 h_w(t) + r_\infty [1 - h_w(t)].
\end{equation}
We find it reasonable to assume that $p(w,r_0,r_\infty) =
p(w)p(r_0)p(r_\infty)$, i.e. that the convergence rates do not depend
upon the initial and final errors.

We may now use these predictions together with some observations to
update $p(w,r_\infty|\xi)$.  More specifically, if $\Pr[r_t = g_w(t
\given r_0, r_\infty) \given r_0, r_\infty, w] = 1$ and we take $m_t$
independent observations $\bz_{t} = (z_t(1), z_t(2), \dotsc,
z_t(m_t))$ of the error with mean $v_t$, the likelihood will be given
by the Bernoulli density
\begin{equation}
  p(\bz_t \given w, r_0, r_\infty) = \left(g_w(t \given r_0, r_\infty)^{v_t}
      [1 - g_w(t \given r_0, r_\infty)]^{1 - v_t}\right)^{m_t}.
\end{equation}
Then it is simple to obtain a posterior density for both $w$ and
$r_\infty$,
\begin{subequations}
  \label{eq:posterior_update}
  \begin{align}
    p(w \given \bz_t)
    &= 
    \frac{p(w)}{p(\bz_t)} \int_0^1 p(\bz_t \given w, r_0, r_\infty = u) \, p(r_\infty = u \given w) \, du\\
    p(r_\infty \given \bz_t)
    &=  \frac{p(r_\infty)}{p(\bz_t)} \int_\Omega p(\bz_t\given w, r_0, r_\infty) \, p(w \given r_\infty) \, dw.
  \end{align}
\end{subequations}
%%where we have made use of the fact that 
%%$p(w, r_\infty) = p(w)p(r_\infty)$

Starting with a prior distribution $p(w \given \xi_0)$ and $p(r_\infty
\given \xi_0)$, we may sequentially update our belief using
\eqref{eq:posterior_update} as follows:
\begin{subequations}
  \begin{align}
    p(w \given \xi_{t+1}) \equiv p(w \given \bz_t, \xi_t)\\
    p(r_\infty \given \xi_{t+1}) \equiv p(r_\infty \given \bz_t, \xi_t).
  \end{align}
\end{subequations}

The realised convergence for a particular training data set may
differ substantially from the expected convergence: the average
convergence curve will be smooth, while any specific instantiation of
it will not be.  More formally, the {\em realised error} given a
specific training dataset is $q_t \equiv \E[R_t \given D^t]$, where
$D^t \sim \CD^t$, while the {\em expected error} given the data
distribution is $r_t \equiv \E[R_t] = \int_{S^t} \E[R_t \given D^t]
\Pr (D_t) \,dD_t$.  The smooth convergence curves that we model would
then correspond to models for $r_t$.

Fortunately, in our case we can estimate a distribution over $r_t$
without having to also estimate a distribution for $q_t$, as this is
integrated out for observations $z \in \{0, 1\}$
\begin{subequations}
  \begin{align}
    p(z \given q_t) &=
    q_t^{z} \, (1-q_t)^{1-z}
    \\
    p(z \given r_t) &=
    \int_0^1 p(z \given q_t) p(q_t = u \given r_t) \, du
%     &=
%    \int_0^1 u^{z} \, (1-u)^{1-z} \, p(q_t = u \given r_t) \, du
%   \\
%     &=
%    \begin{cases}
%      \int_0^1 (1 - u) \, p(q_t = u \given r_t) \, du, & z =0\\
%    \int_0^1 u \, p(q_t = u \given r_t) \, du, & z =1
%    \end{cases}
%    \\
%     &=
%    \begin{cases}
%      1 - \E[R_t \given r_t], &z = 0\\
%     \E[R_t \given r_t] , & z =1
%    \end{cases}
%    \\
    = r_t^{z} \, (1 - r_t) ^{1 - z}.
  \end{align}
\end{subequations}
%%since by definition $r_t = \E[R_t \given r_t]$.

\subsubsection{Step 5.  Deciding whether to Stop}
We may now use the distribution over the models to predict the error
should we choose to add $k$ more examples.  This is simply
\[
\E[R_{t+k} \given \xi_t] = 
\int_0^1 \int_{\Omega} g_w(t+k \given r_0, r_\infty) p(w \given \xi_t) 
     p(r_\infty \given \xi_t) \, dw \, dr_\infty.
\]
The calculation required for step 8 of {OBSV} follows trivially.

\begin{algorithm}
  Given a dataset $D$ with examples in $N_c$ classes and any learning
  algorithm $F$, initial beliefs $\Pr(w \given \xi_0)$ and
  $\Pr(r_\infty \given \xi_0)$ and a method for updating them, and
  additionally a known query cost $\gamma$ for discovering the class
  label $y_i \in [1, \dotsc, n]$ of example $i \in D$,
  \begin{algorithmic}[1]        
    \STATE Split $D$ into $D_A, D_R$.
    \STATE $r_0 = 1 - 1/N_c$.
    \STATE Initialise the classifier $f$.
    \FOR{$t = 1, 2,\dotsc$}     
    \STATE Sample $i \in D_R$ without replacement and observe $f(x_i), y_i$ to calculate $v_t$.   
    \STATE  Calculate $\Pr(w,r_\infty \given \xi_{t}) \equiv \Pr(w,r_\infty \given v_t, \xi_{t-1})$.                       
    \STATE If $D_A \neq \emptyset$, set $k=2$, otherwise $k=1$.
    \IF {$\E[R_{t+k} \given \xi_t] + k\gamma < \E[R_t \given \xi_t]$}
    \STATE Exit.
    \ENDIF     
    \STATE If $D_A \neq \emptyset$, use $F$ to query a new example $j \in D_A$ without replacement, $D_T \leftarrow D_T \cup j$. 
    \STATE $D_T \leftarrow D_T \cup i$, $f \leftarrow F(D_T)$.                               
                
    \ENDFOR
  \end{algorithmic}
  \caption{OBSV, a specific instantiation of the bounded stopping algorithm.}
  \label{al:obsv}
\end{algorithm}

\subsubsection{Specifics of the Model}
\label{sec:estimates}
What remains unspecified is the set of convergence curves that will be
employed.  We shall make use of curves related to common theoretical
convergence results.  It is worthwhile to keep in mind that we simply
aim to find the combination of the available estimates that gives the
best predictions.  While none of the estimates might be particularly
accurate, we expect to obtain reasonable stopping times when they are
optimally combined in the manner described in the previous section.
Ultimately, we expect to end up with a fairly narrow distribution over
the possible convergence curves.

One of the weakest convergence results~\cite{alt:kaariainen:2006} is
for sample complexity of order $\CO(1/\epsilon_t^2)$, which corresponds to
the convergence curve
\begin{equation}
  \label{eq:quadratic_estimate}
  h_{q}(t) = \sqrt{\frac{\kappa}{t + \kappa}}, \quad \kappa \geq 1
\end{equation}
Another common type is for sample complexity of order
$\CO(1/\epsilon_t)$, which corresponds to the curve
\begin{equation}
  \label{eq:geometric_estimate}
  h_{g}(t) = \frac{\lambda}{t + \lambda}, \quad \lambda \geq 1
\end{equation}
A final possibility is that the error decreases exponentially fast.
This is theoretically possible in some cases, as was proven in
\cite{icml:Balcan:2006}. The resulting sample complexity of order
$\CO(\log(1/\epsilon_t))$ corresponds to the convergence curve
\begin{equation}
  \label{eq:exponential_estimate}
  h_{exp}(t) = \beta^t, \quad \beta \in (0, 1).
\end{equation}
Since we do not know what appropriate values of the constants $\beta$,
$\lambda$ and $\kappa$, are, we will model this uncertainty as an
additional distribution over them, i.e.  $p(\beta \given \xi_t)$.
This would be updated together with the rest of our belief
distribution and could be done in some cases analytically.  In this
paper however we consider approximating the continuous densities by a
sufficiently large set of models, one for each possible value of the
unknown constants.

%% [NOTE: Plot how the best stopping time changes over time].

\begin{figure}[htb]
  \begin{center}
    \includegraphics[width=0.9\textwidth]{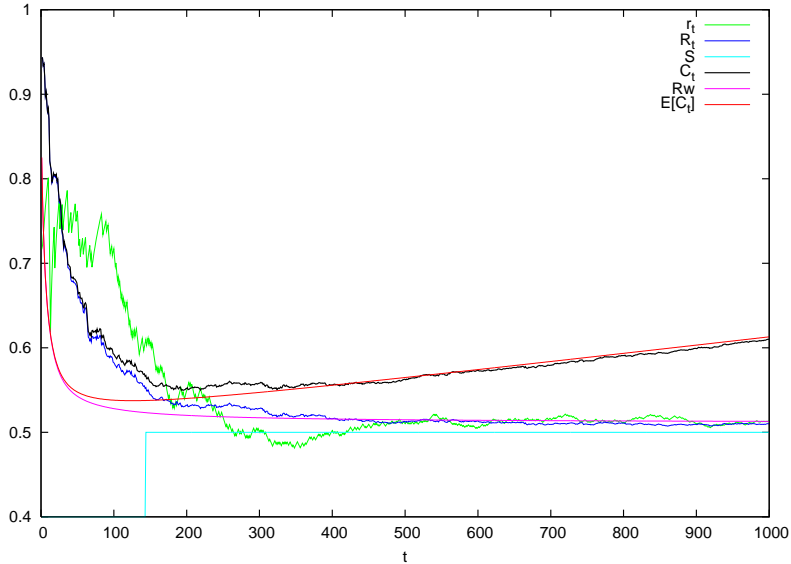}
    \end{center}
    \caption{Illustration of the estimated error on a 10-class problem
      with a cost per label of $\gamma = 0.001$. On the vertical axis,
      $r_t$ is the {\bf history} of the predicted generalisation error,
    i.e $\E[r_t \given \xi_{t-1}]$, while $R_t$ is the {\bf
      generalisation error} measured on a test-set of size 10,000 and
    $C_t$ is the corresponding actual {\bf cost}. Finally, $R_w$ and
    $E[C_t]$ are the final {\bf estimated} convergence and cost {\em
      curves} given all the observations.  The stopping time is
    indicated by $S$, which equals $0.5$ whenever \refalg{al:obsv}
    decides to stop and $t$ is the number of iterations.}
    \label{fig:convergence_illustration}
\end{figure}
As a simple illustration, we examined the performance of the
estimation and the stopping criterion in a simple classification
problem with data of 10 classes, each with an equivariant Gaussian
distribution in an 8-dimensional space.  Each unknown point was simply
classified as having the label closest to the empirical mean of the
observations for each class.  Examples were always chosen randomly.

As can be seen in \reffig{fig:convergence_illustration}, at the
initial stages the estimates are inaccurate.  This is because of two
reasons:
\begin{inparaenum}[(a)]
\item The distribution over convergence rates is initially dominated
  by the prior. As more data is accumulated, there is better evidence
  for what the final error will be.
\item As we mentioned in the discussion of step 6, the realised
  convergence curve is much more random than the expected convergence
  curve which is actually modelled.  However, as the number of
  examples approaches infinity, the expected and realised errors
  converge.
\end{inparaenum}
The stopping time for \refalg{al:obsv} (indicated by $S$) is
nevertheless relatively close to the optimal stopping time, as $C_t$
appears to be minimised near 200.  The following section presents a
more extensive evaluation of this stopping algorithm.

\section{Experimental Evaluation}
\label{sec:experiments}
The main purpose of this section is to evaluate the performance of the
OBSV stopping algorithm.  This is done by examining its cost and
stopping time when compared to the optimal stopping time.  Another aim
of the experimental evaluation was to see whether mixed sampling
strategies have an advantage compared to random sampling strategies
with respect to the cost, when the stopping time is decided using a
stopping algorithm that takes into account the labelling cost.
Following~\cite{Bengio:Expected:2005}, we plot performance curves for
a range of values of $\gamma$, utilising multiple runs of
cross-validation in order to assess the sensitivity of the results to
the data.  For each run, we split the data into a training set $D$ and
test set $D_E$, the training set itself being split into random and
mixed sampling sets whenever appropriate.

More specifically, we compare the OBSV algorithm with the {\bf oracle}
stopping time.  The latter is defined simply as the stopping time
minimising the cost as this is measured on the independent test set
for that particular run.  We also compare {\bf random} sampling with
{\bf mixed} sampling.  In random sampling, we simply query unlabelled
examples without replacement.  For the mixed sampling procedure, we
actively query an additional label for the example from $D_A$ closest
to the decision boundary of the current classifier, also without
replacement. This strategy relies on the assumption that those labels
are most informative
~\cite{icml:abe:1998},~\cite{icml:Melville:2004},~\cite{ecml:Korner:2006}
and thus convergence will be faster.  Stopping times and cost ratio
curves are shown for a set of $\gamma$ values, for costs as defined
in~\eqref{eq:expected_cost}.  These values of $\gamma$ are also used
as input to the stopping algorithm.  The ratios are used both to
compare stopping algorithms (OBSV versus the oracle) and sampling
strategies (random sampling, where $D_A=\emptyset$, and mixed
sampling, with $|D_A|=|D_R|$).  Average test error curves are also
plotted for reference.

%{\bf %[CD: Write something about the ranges of $\gamma$: i.e. with the
%  given prior, OBSV never even takes one observation if $\gamma >
%  10^{-2}$.  Something like 'as you can see, the prior is not
%  optimal... blah - nevertheless..'. 
%  %
%Also note
%  that we always stop a bit faster than the oracle stopping algorithm.
%  
%An alternative algorithm would stop if the estimated error reduction
%  in the PREVIOUS time step was also smaller than $\gamma$.  Future
%  work?]}

For the experiments we used two data sets from the UCI
repository\footnote{http://mlearn.ics.uci.edu/MLRepository.html}: the
Wisconsin breast cancer data set ({\wdbc}) with 569 examples and the
spambase database ({\spam}) with 4601 examples. We evaluated {\wdbc}
and {\spam} using 5 and 3 randomised runs of 3-fold stratified
cross-validation respectively.  The classifier used was
AdaBoost~\cite{JCSS:Freund+Schapire:1997} with 100 decision stumps as
base hypotheses. Hence we obtain a total of 15 runs for {\wdbc} and 9
for {\spam}.  We ran experiments for values of $\gamma \in\{9\cdot
10^{-k}, 8\cdot 10^{-k}, \dots, 1\cdot 10^{-k}\}$, with $k=1,\dots,
7$, and $\gamma=0$.  For every algorithm and each value of $\gamma$ we
obtain a different stopping time $t_{\gamma}$ for each run.  We then
calculate $v_e(\gamma,F,t_\gamma)$ as given in
\eqref{eq:expected_cost} on the corresponding test set of the run.  By
examining the averages and extreme values over all runs we are able to
estimate the sensitivity of the results to the data.

\begin{figure*}[h]
\begin{center} 
  \subfigure[]{
    \includegraphics[width=0.47\textwidth]{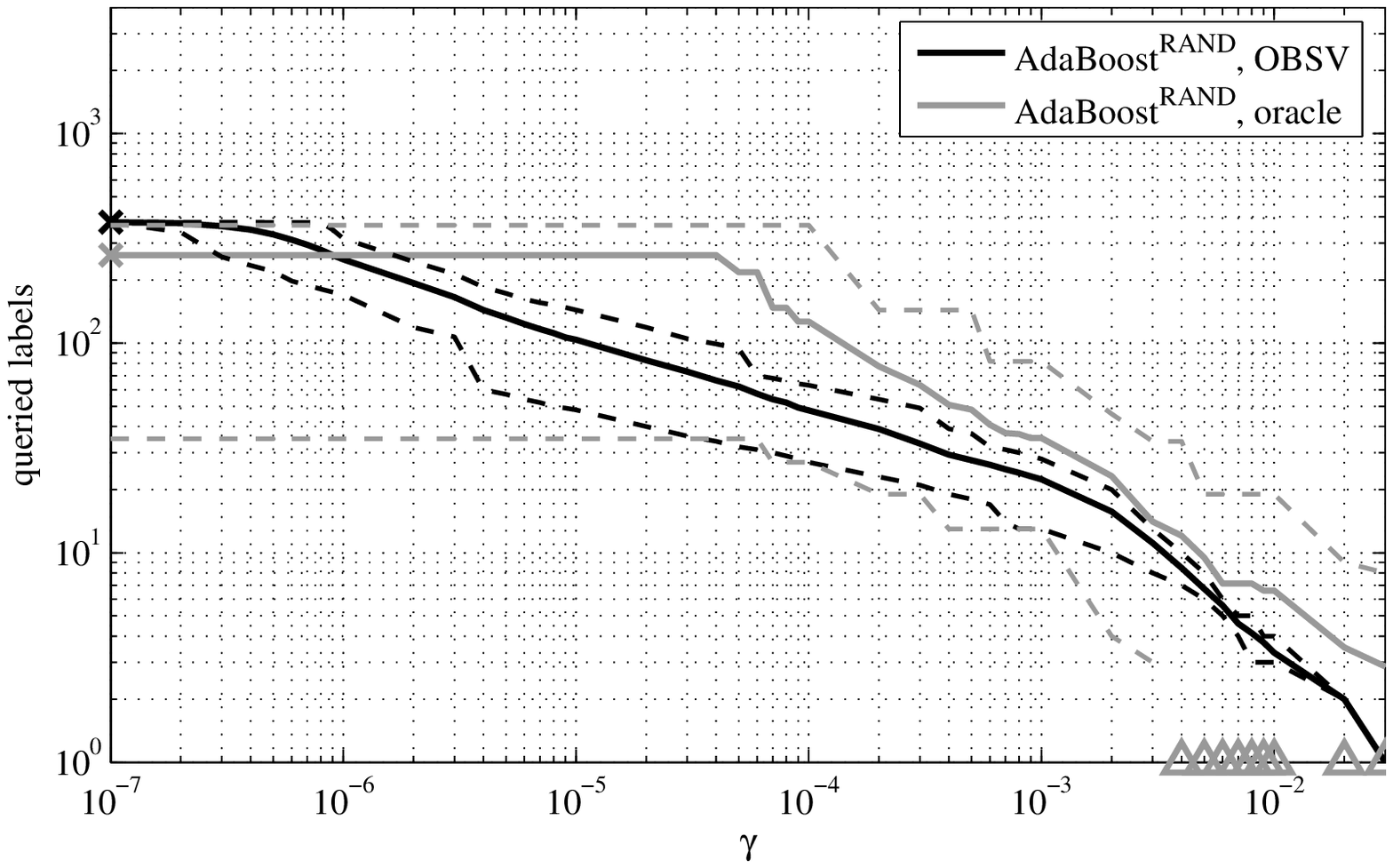}
	\label{fig:wdbc_TStop_random_obsv_oracle}
	}
   \subfigure[]{
    \includegraphics[width=0.47\textwidth]{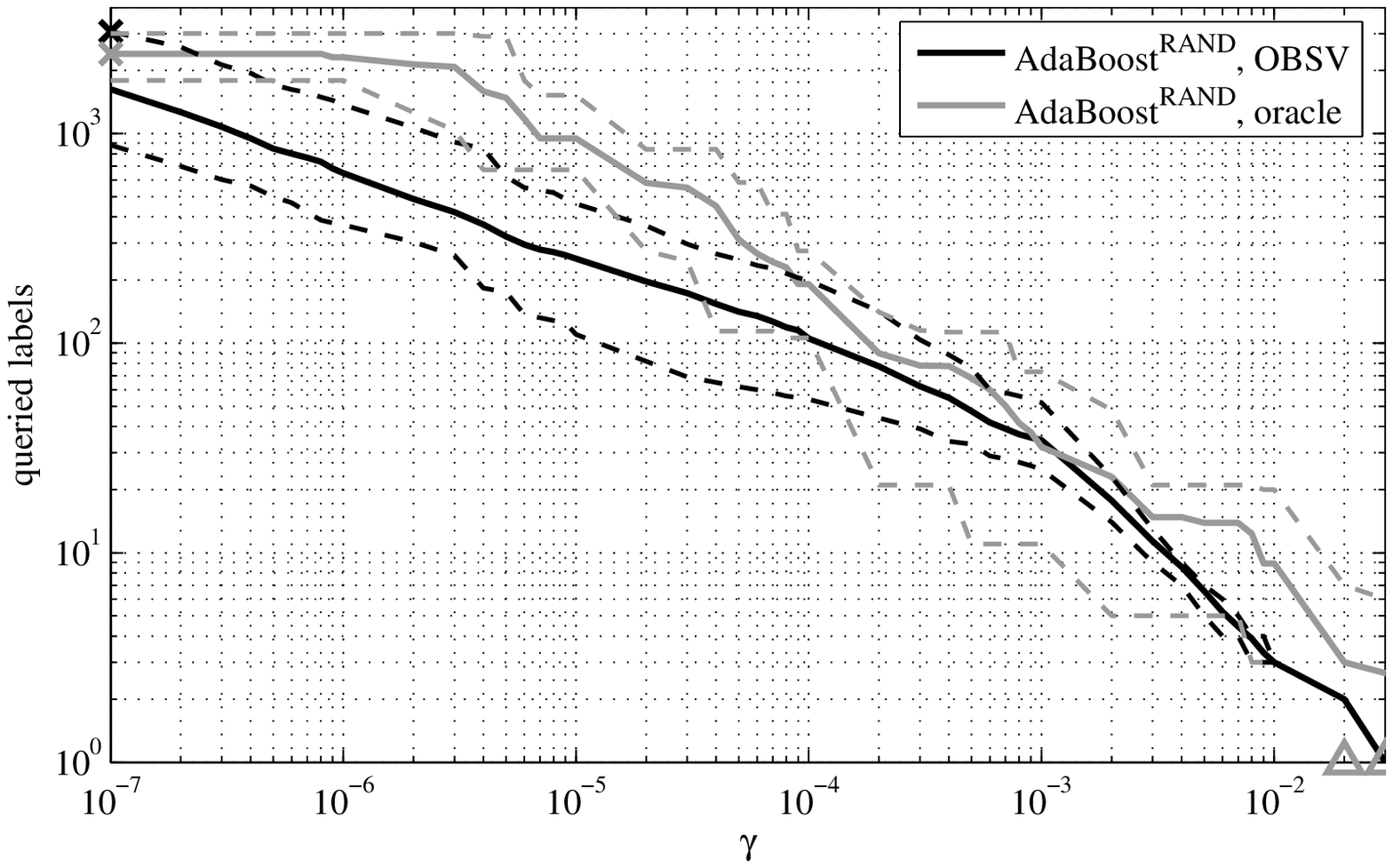}
	\label{fig:spam_TStop_random_obsv_oracle}
  }
  \subfigure[]{
    \includegraphics[width=0.47\textwidth]{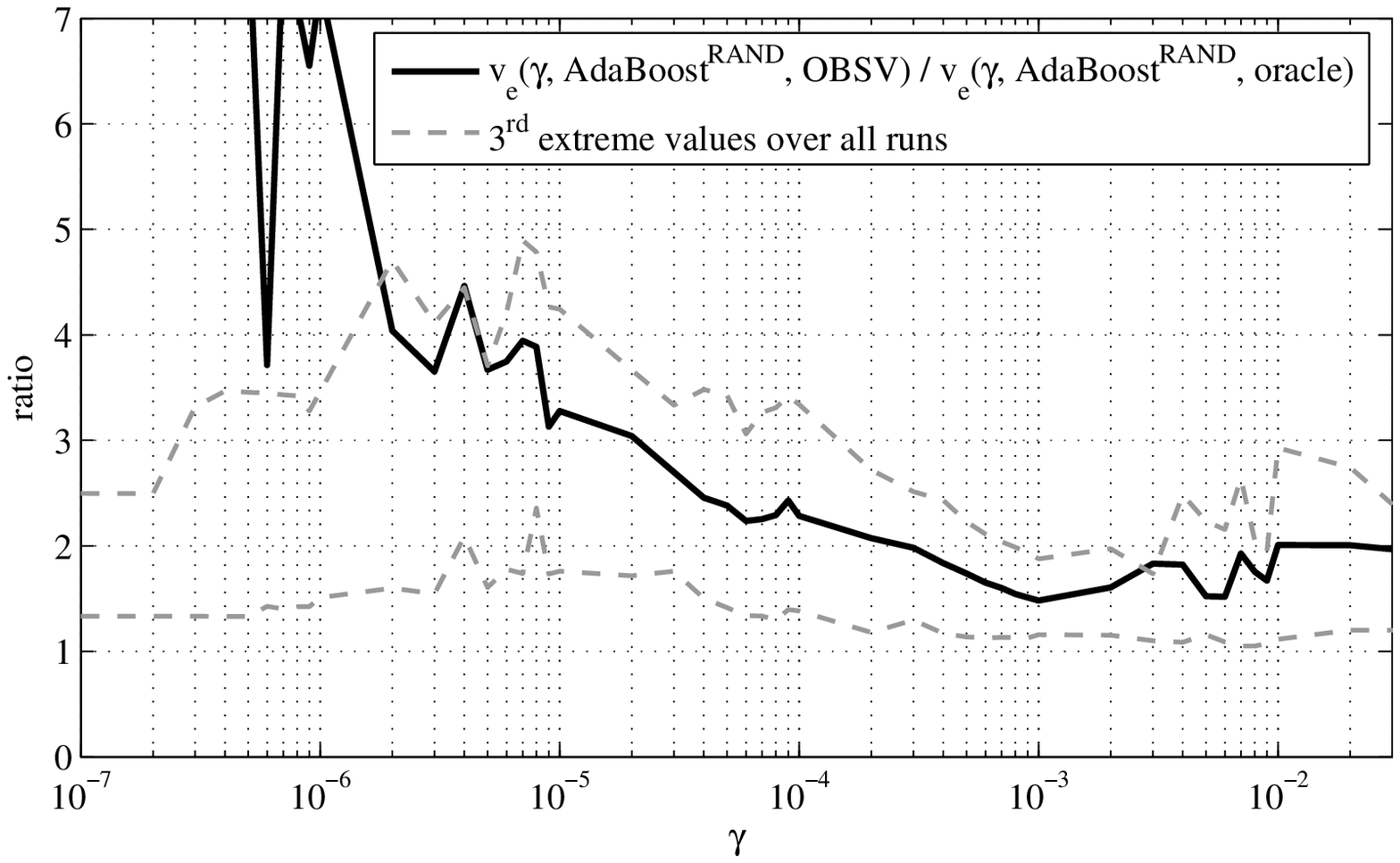}
	\label{fig:wdbc_CostRatio_random_obsv_oracle}
  }     
  \subfigure[]{
    \includegraphics[width=0.47\textwidth]{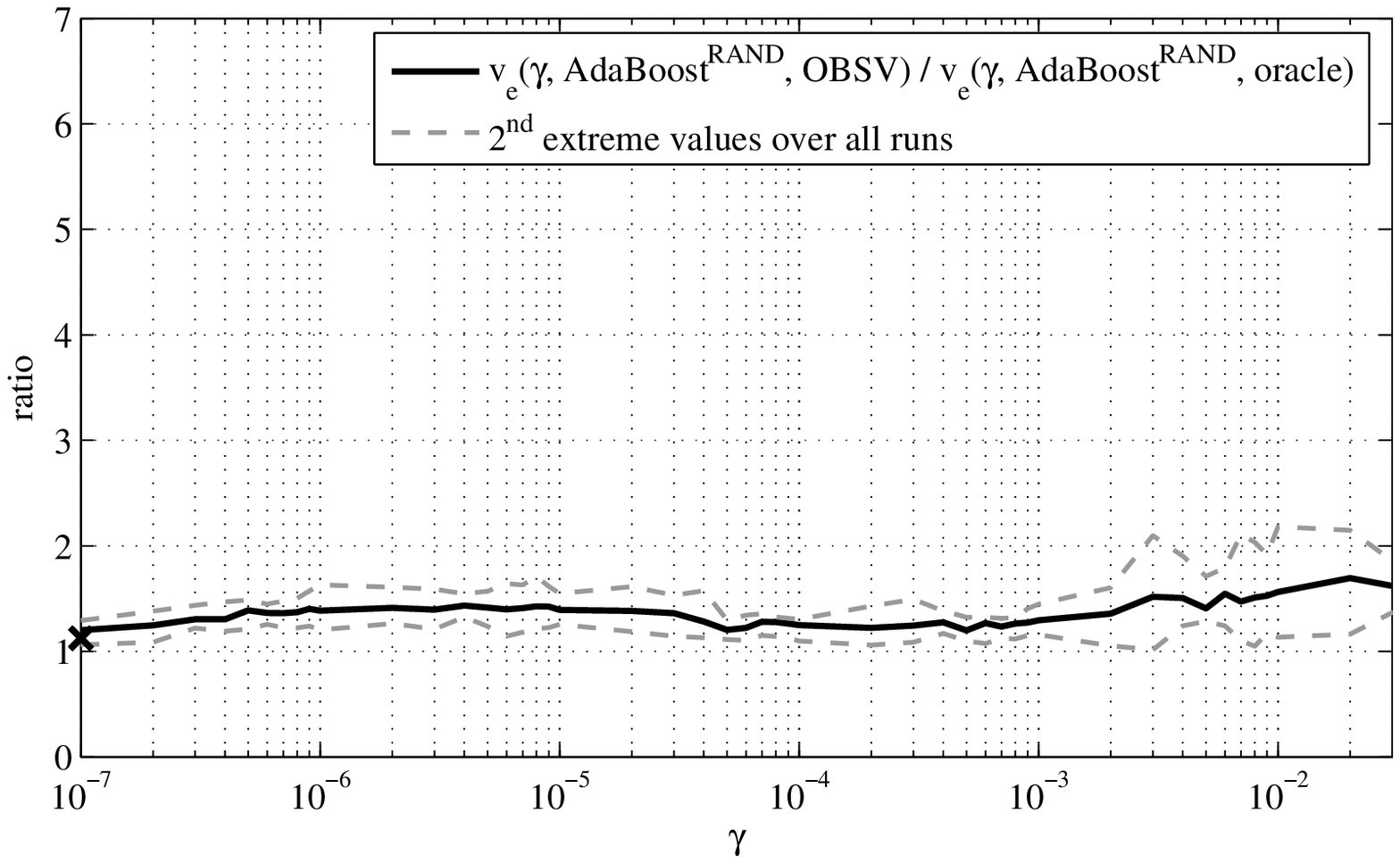}
	\label{fig:spam_CostRatio_random_obsv_oracle}
  }  
  \end{center}
  \caption{Results for \textbf{random sampling} on the {\wdbc} (left column) 
    and the {\spam} data (right column) as obtained from the 15 ({\wdbc}) and 9 
    ({\spam}) runs of AdaBoost with 100 decision stumps. The first row
    \subref{fig:wdbc_TStop_random_obsv_oracle},
    \subref{fig:spam_TStop_mixed_obsv_oracle},
    plots the average stopping times from OBSV and the oracle as a function 
    of the labelling cost $\gamma$. For each $\gamma$ the extreme values 
    from all runs are denoted by the dashed lines. The second
    row, \subref{fig:wdbc_CostRatio_random_obsv_oracle},
    \subref{fig:spam_CostRatio_random_obsv_oracle}, shows the
    corresponding average ratio in $v_e$ over all runs between 
    OBSV and the oracle, where for each $\gamma$ the $3^{rd}$~({\wdbc}) 
    / $2^{nd}$~({\spam}) extreme values from all runs are denoted by 
    the dashed lines. Note a zero value on a logarithmic scale is denoted 
    by a cross or by a triangle. Note for {\wdbc} and smaller values of $\gamma$ 
    the average ratio in $v_e$ sometimes exceeds the denoted extreme 
    values due to a zero test error occurred in one run.}
  \label{fig:wdbc_spam_random}
\end{figure*}

The results comparing the oracle with OBSV for the {\bf random}
sampling strategy\footnote{The corresponding average test errors can
  be seen in
  \reffigseptwo{fig:wdbc_CVErrLog_random_mixed}{fig:spam_CVErrLog_random_mixed}.}
are shown in \reffig{fig:wdbc_spam_random}. In
\reffigseptwo{fig:wdbc_TStop_random_obsv_oracle}{fig:spam_TStop_random_obsv_oracle}
it can be seen that the stopping times of OBSV and the oracle increase
at a similar rate. However, although OBSV is reasonably close, on
average it regularly stops earlier.  This may be due to a number of
reasons.  For example, due to the prior, OBSV stops immediately when
$\gamma>3\cdot10^{-2}$.  At the other extreme, when $\gamma\rightarrow
0$ the cost becomes the test error and therefore the oracle always
stops at latest at the minimum test error\footnote{This is obtained
  after about $260$ labels on {\wdbc} and $2400$ labels on {\spam}}.
This is due to the stochastic nature of the realised error curve,
which cannot be modelled; there, the perfect information that the
oracle enjoys accounts for most of the performance difference.  As
shown in
\reffigseptwo{fig:wdbc_CostRatio_random_obsv_oracle}{fig:spam_CostRatio_random_obsv_oracle},
the extra cost induced by using OBSV instead of the oracle is bounded
from above for most of the runs by factors of $2$ to $5$ for {\wdbc}
and around $0.5$ for {\spam}.  The rather higher difference on {\wdbc}
is partially a result of the small dataset.  Since we can only measure
an error in quanta of $1/|D_E|$, any actual performance gain lower
than this will be unobservable.  This explains why the number of
examples queried by the oracle becomes constant for a value of
$\gamma$ smaller than this threshold.  Finally, this fact also
partially explains the greater variation of the oracle's stopping time
in the smaller dataset.  We expect that with larger test sets, the
oracle's behaviour would be smoother.

\begin{figure*}[h]
\begin{center} 
  \subfigure[\hspace{0.01cm}]{
    \includegraphics[width=0.47\textwidth]{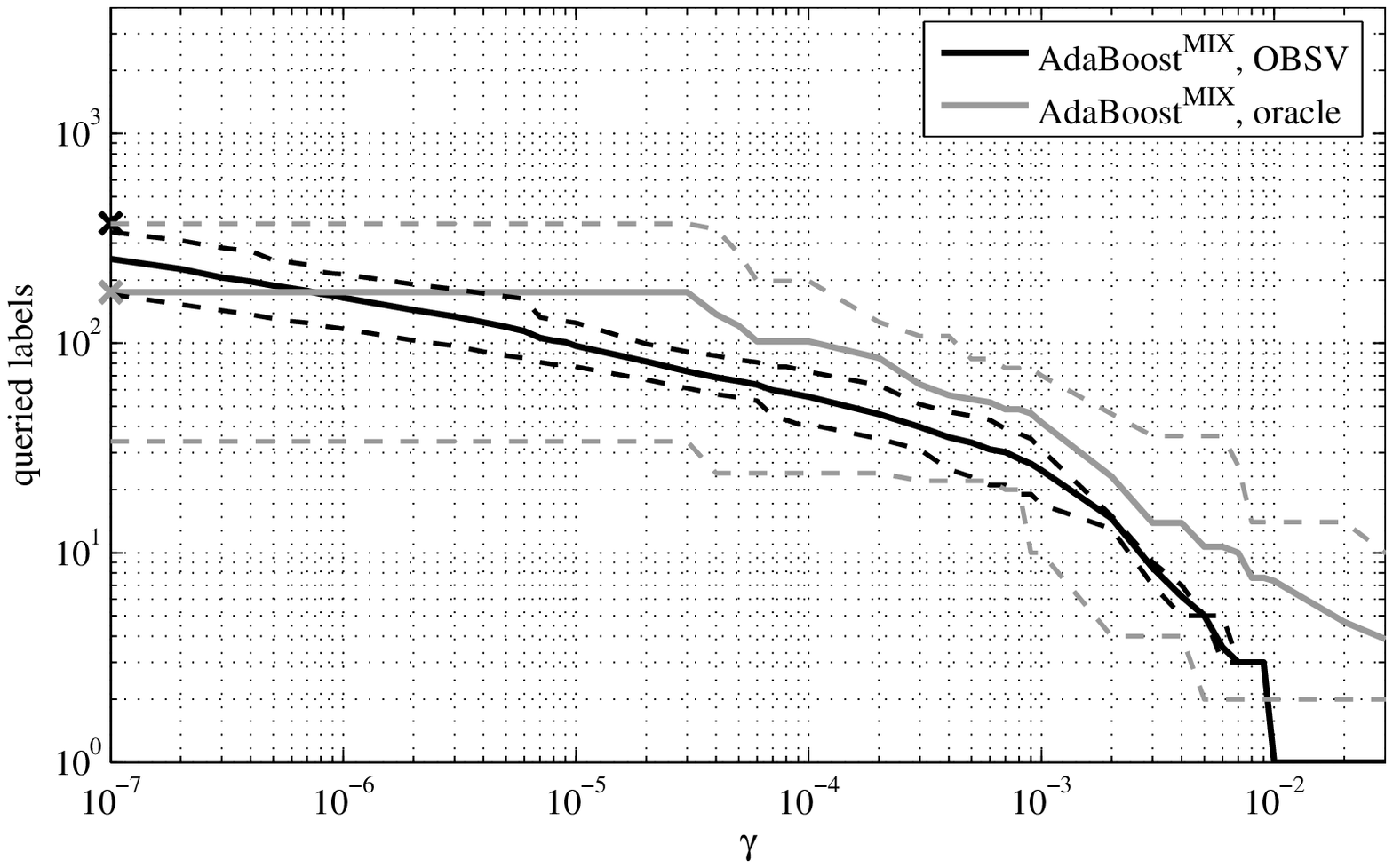}
	\label{fig:wdbc_TStop_mixed_obsv_oracle}
	}
   \subfigure[\hspace{0.01cm}]{
    \includegraphics[width=0.47\textwidth]{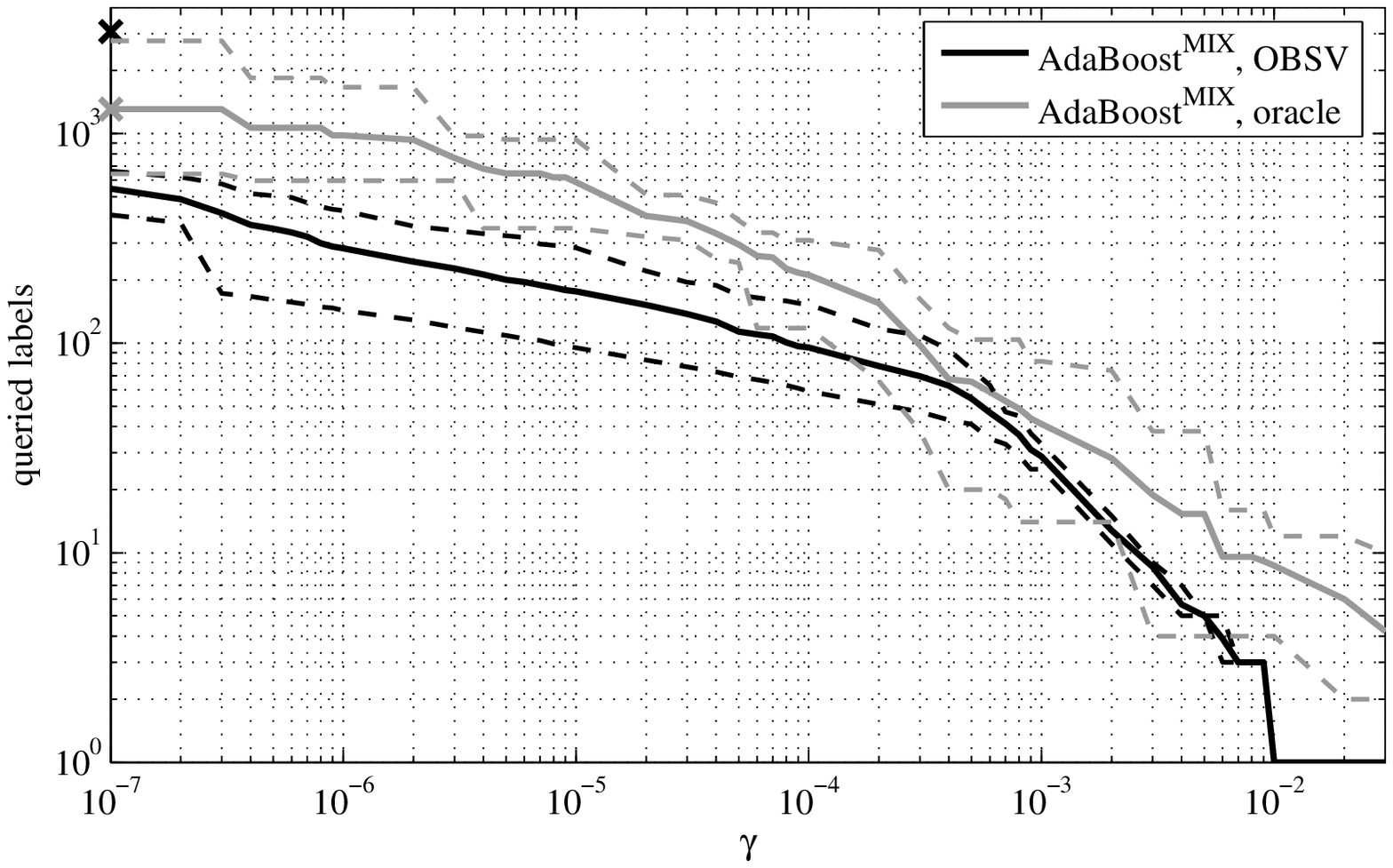}
	\label{fig:spam_TStop_mixed_obsv_oracle}
  }
  \subfigure[\hspace{0.01cm}]{
    \includegraphics[width=0.47\textwidth]{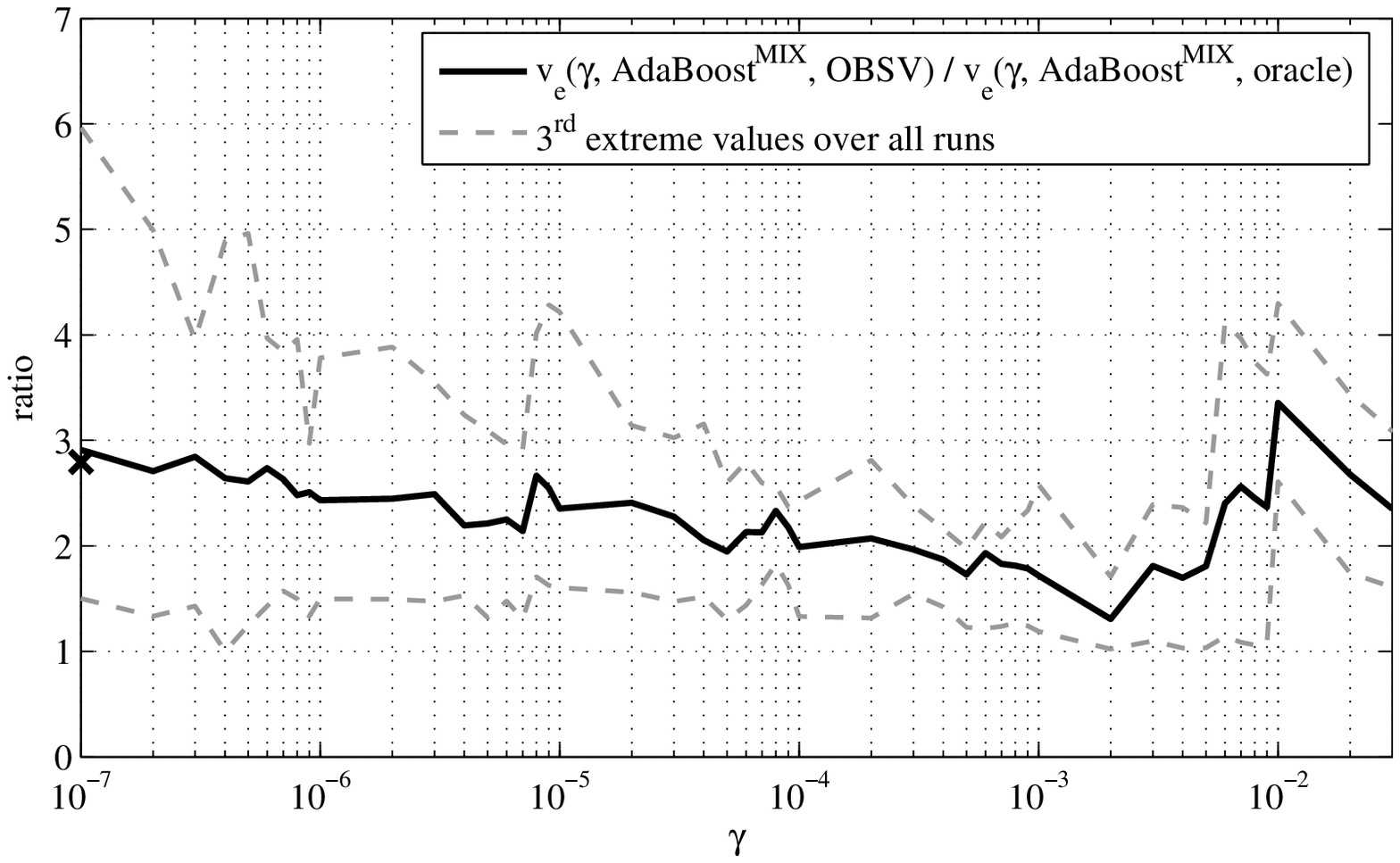}
	\label{fig:wdbc_CostRatio_mixed_obsv_oracle}
  }     
  \subfigure[\hspace{0.01cm}]{
    \includegraphics[width=0.47\textwidth]{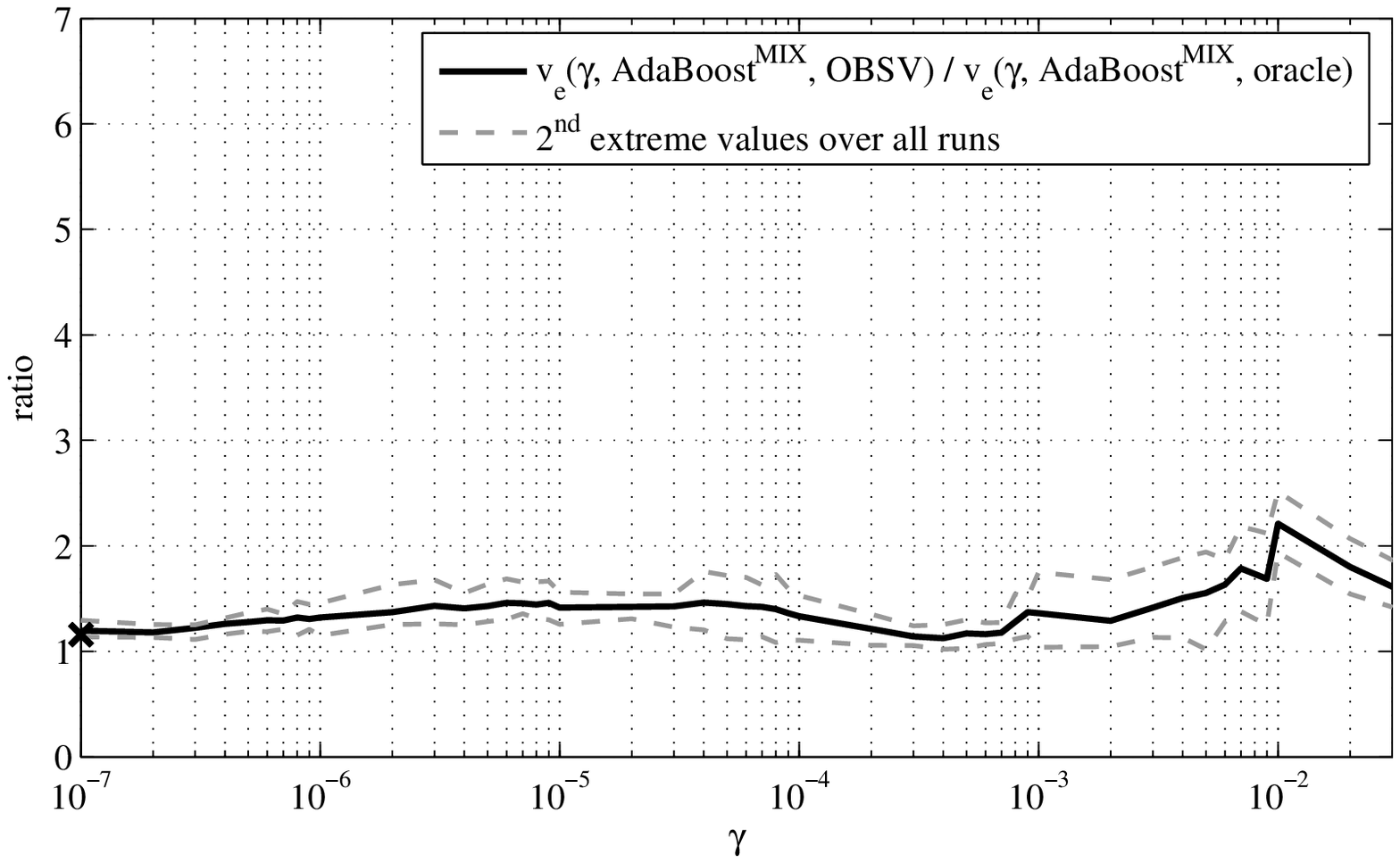}
	\label{fig:spam_CostRatio_mixed_obsv_oracle}
  }  
  \end{center}
  \caption{Results for \textbf{mixed sampling} on the {\wdbc} (left column) 
    and the {\spam} data (right column) as obtained from the 15 ({\wdbc}) and 
    9 ({\spam}) runs of AdaBoost with 100 decision stumps. The first row
    \subref{fig:wdbc_TStop_mixed_obsv_oracle},
    \subref{fig:spam_TStop_mixed_obsv_oracle},
    plots the average stopping times from OBSV and the oracle as a function 
    of the labelling cost $\gamma$. For each $\gamma$ the extreme values 
    from all runs are denoted by the dashed lines. The second
    row, \subref{fig:wdbc_CostRatio_mixed_obsv_oracle},
    \subref{fig:spam_CostRatio_mixed_obsv_oracle}, shows the
    corresponding average ratio in $v_e$ over all runs between 
    OBSV and the oracle, where for each $\gamma$ the $3^{rd}$~({\wdbc}) 
    / $2^{nd}$~({\spam}) extreme values from all runs are denoted by 
    the dashed lines. Note a zero value on a logarithmic scale is denoted 
    by a cross.
}
  \label{fig:wdbc_spam_mixed}
\end{figure*}

\begin{figure*}[h]
\begin{center} 
  \subfigure[\hspace{0.01cm}]{
    \includegraphics[width=0.47\textwidth]{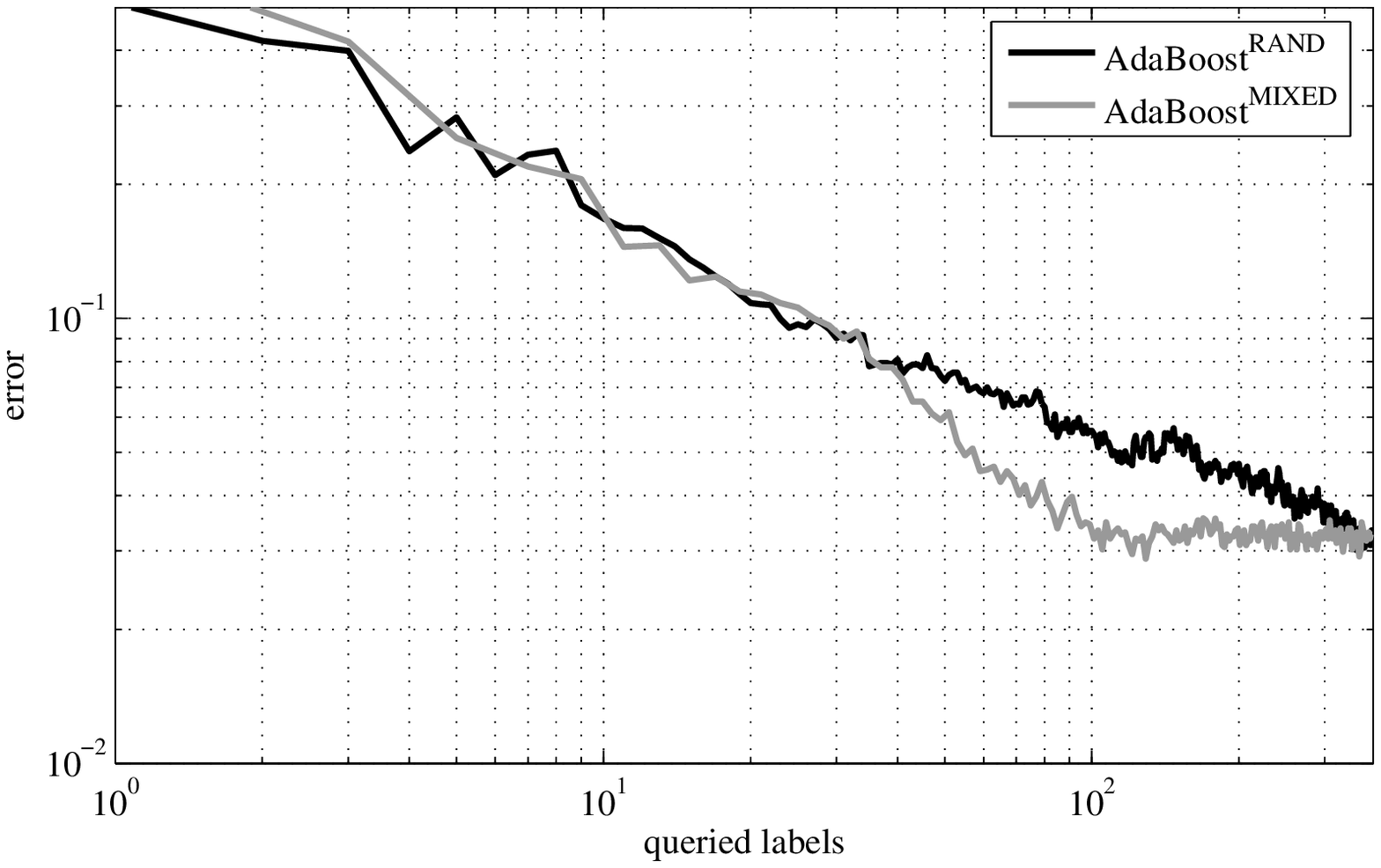}
	\label{fig:wdbc_CVErrLog_random_mixed}
  }
  \subfigure[\hspace{0.01cm}]{
    \includegraphics[width=0.47\textwidth]{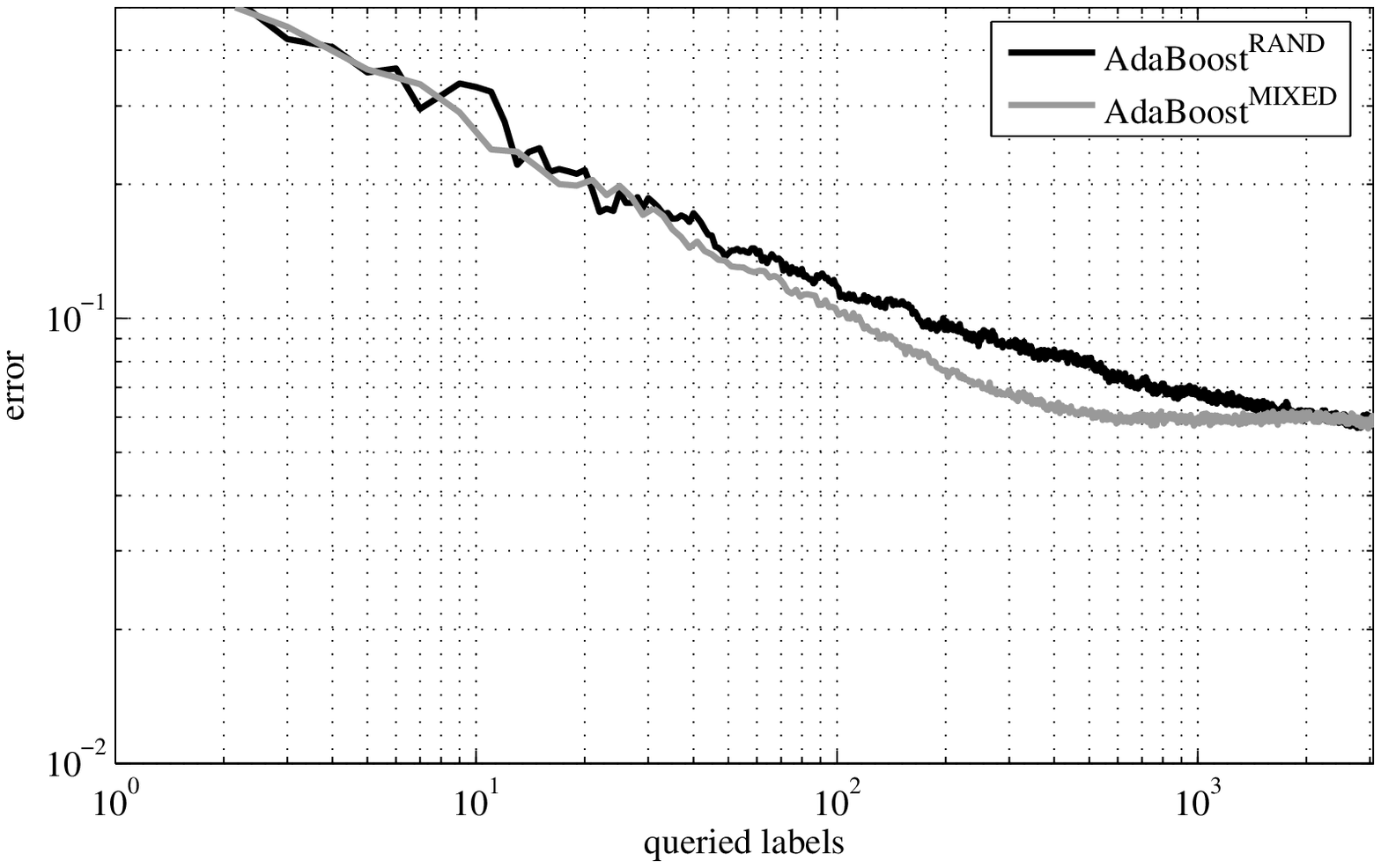}
	\label{fig:spam_CVErrLog_random_mixed}
  }
  \subfigure[\hspace{0.01cm}]{
    \includegraphics[width=0.47\textwidth]{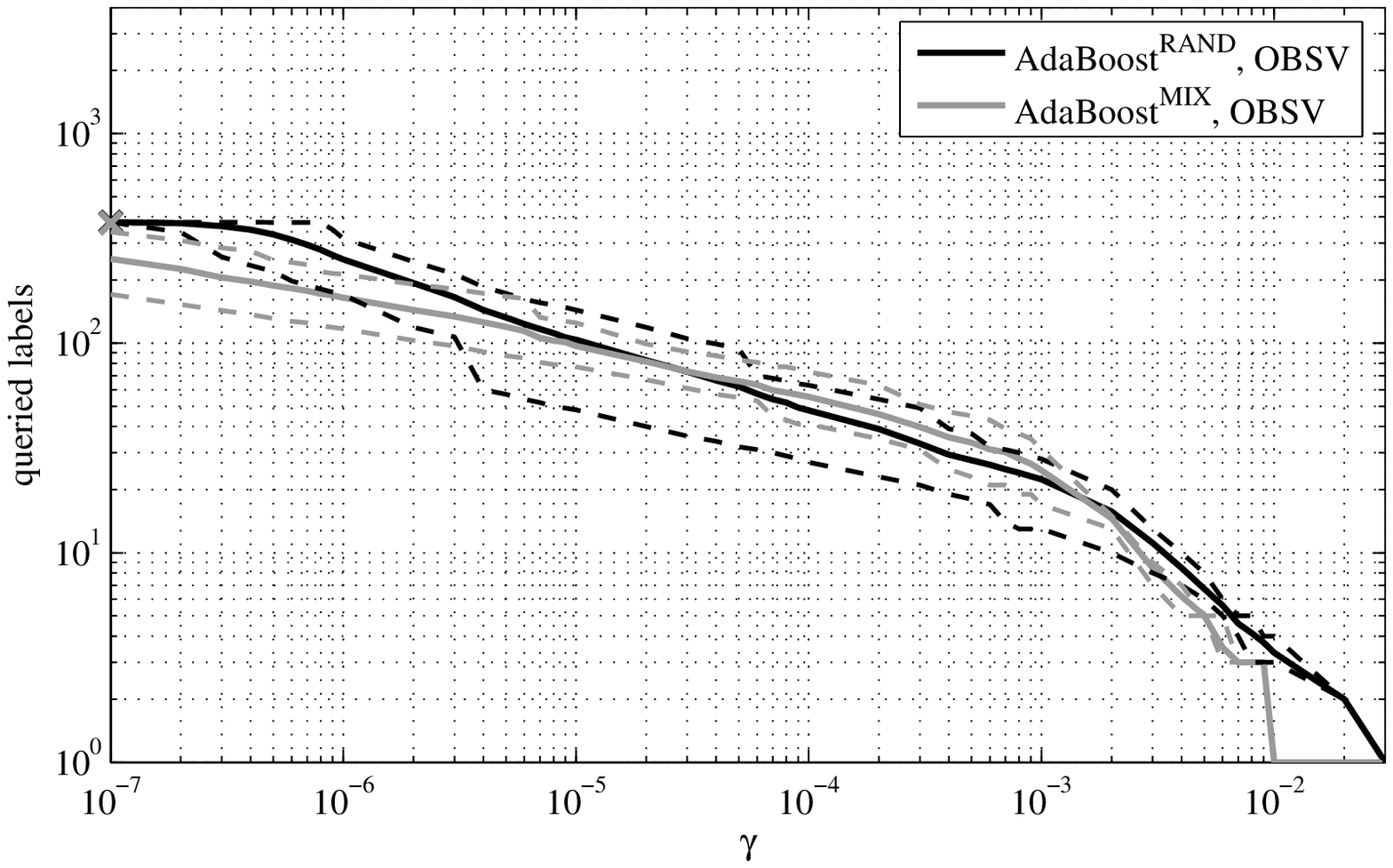}
	\label{fig:wdbc_TStop_random_mixed_obsv}
	}
   \subfigure[\hspace{0.01cm}]{
    \includegraphics[width=0.47\textwidth]{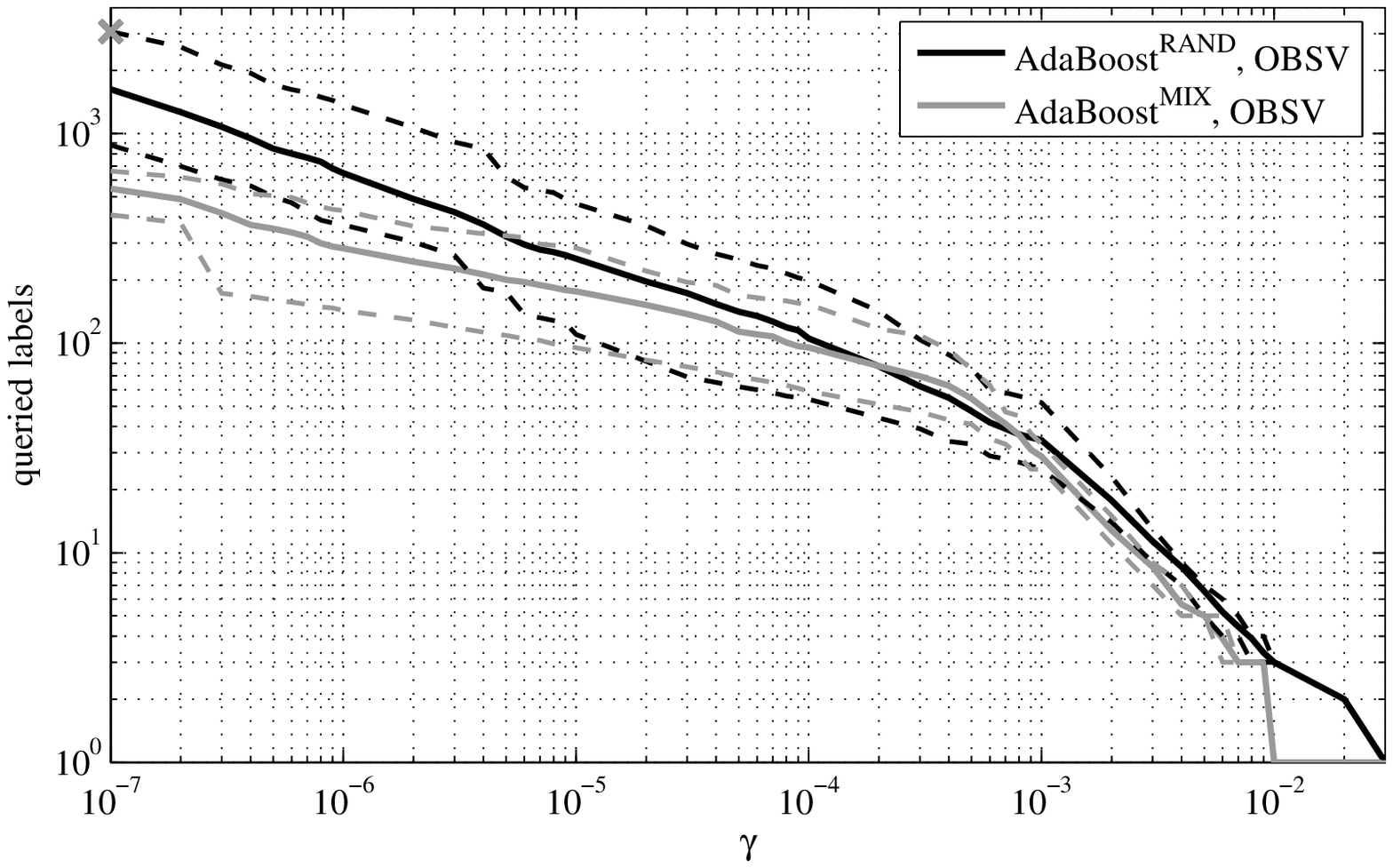}
	\label{fig:spam_TStop_random_mixed_obsv}
  }
  \subfigure[\hspace{0.01cm}]{
    \includegraphics[width=0.47\textwidth]{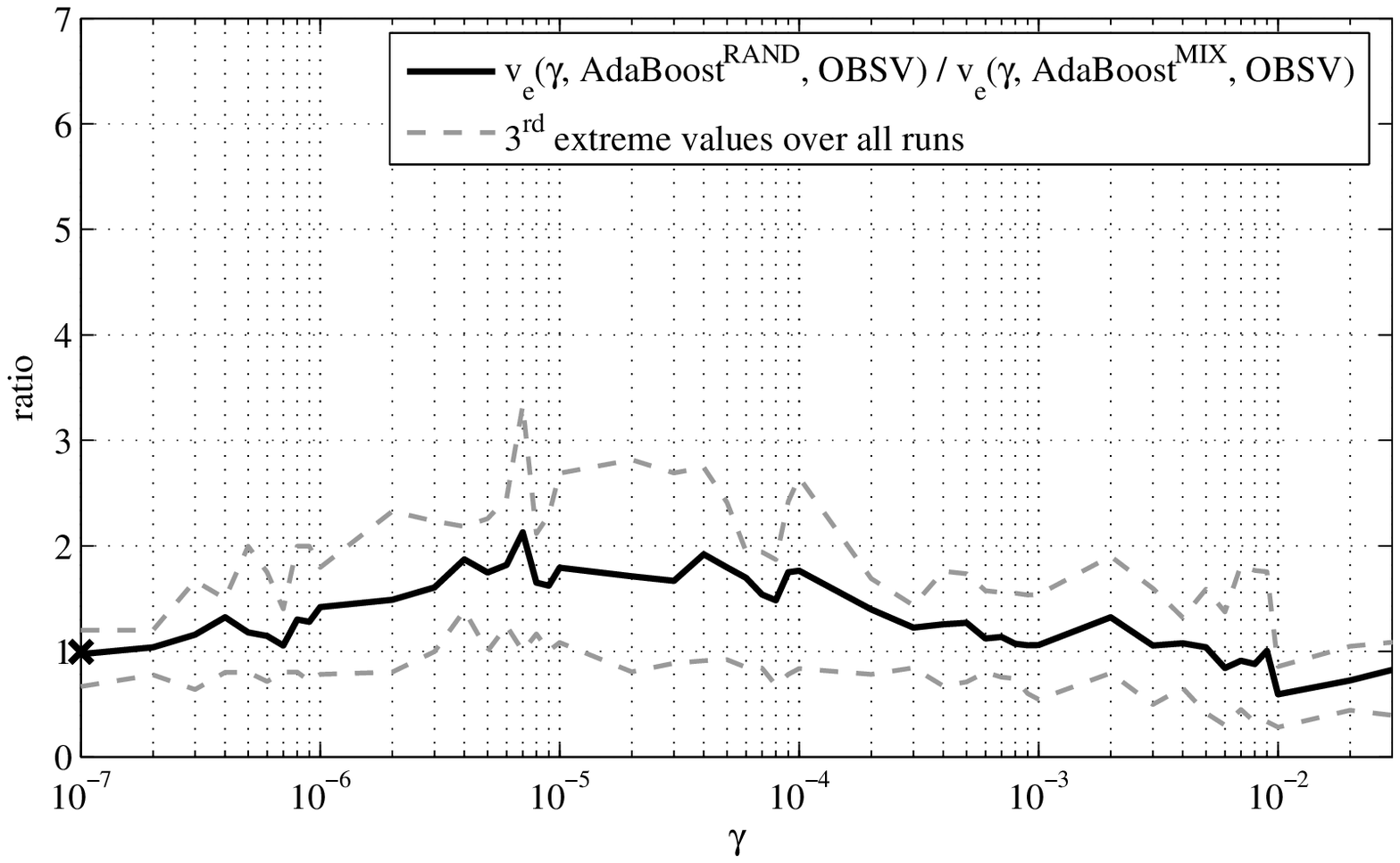}
	\label{fig:wdbc_CostRatio_random_mixed_obsv}
  }     
  \subfigure[\hspace{0.01cm}]{
    \includegraphics[width=0.47\textwidth]{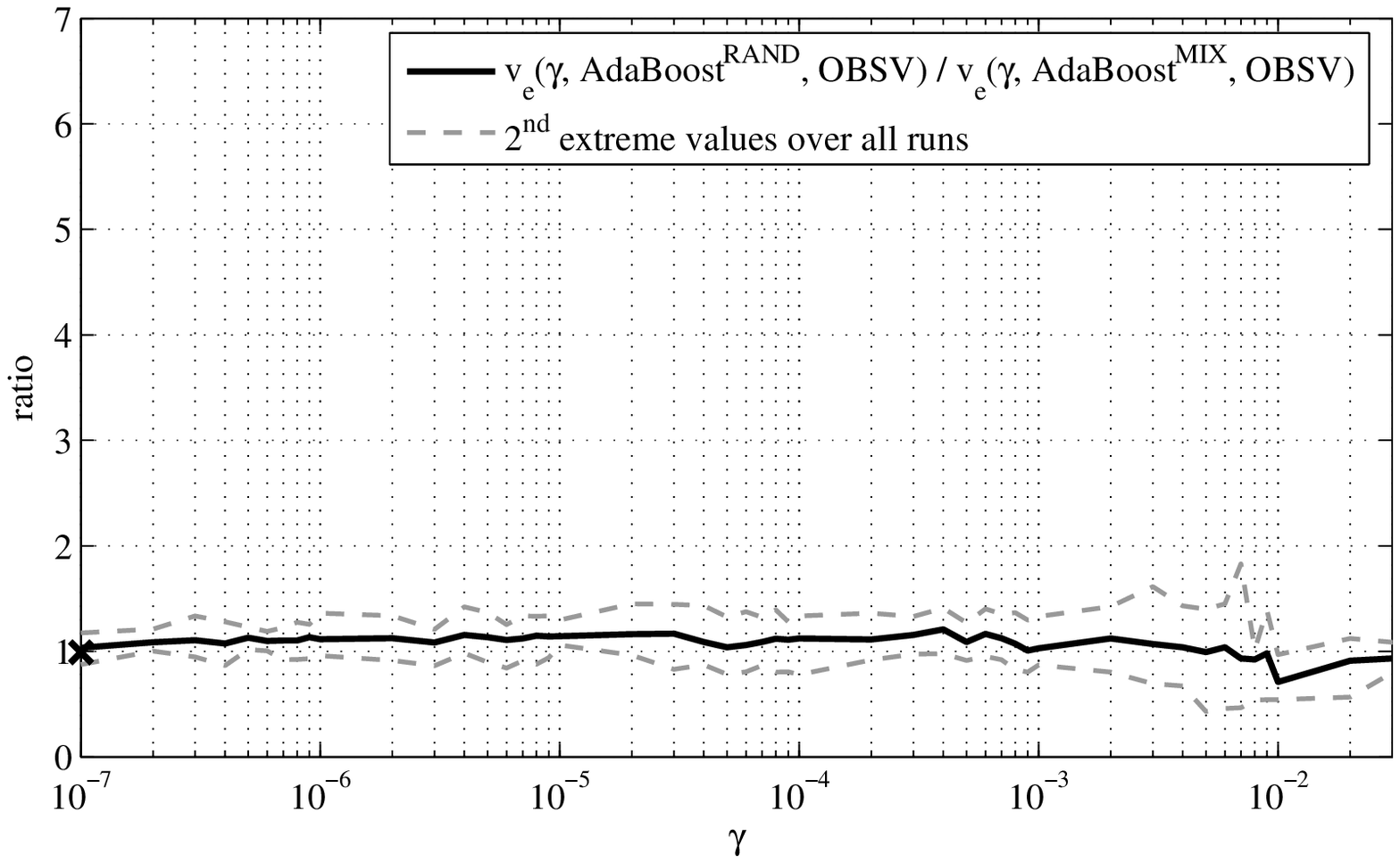}
	\label{fig:spam_CostRatio_random_mixed_obsv}
  }  
  \end{center}
  \caption{Results comparing random ({\bf RAND}) and mixed ({\bf MIX})
    sampling on the {\wdbc} (left column) and the {\spam} data (right
    column) as obtained from the 15 ({\wdbc}) and 9 ({\spam}) runs of
    AdaBoost with 100 decision stumps.  The first row
    \subref{fig:wdbc_CVErrLog_random_mixed},
    \subref{fig:spam_CVErrLog_random_mixed}, shows the test error of
    each sampling strategy averaged over all runs.  The second row
    \subref{fig:wdbc_TStop_mixed_obsv_oracle},
    \subref{fig:spam_TStop_mixed_obsv_oracle}, plots the average
    stopping times from OBSV and the oracle as a function of the
    labelling cost $\gamma$. For each $\gamma$ the extreme values from
    all runs are denoted by the dashed lines.  The third row,
    \subref{fig:wdbc_CostRatio_mixed_obsv_oracle},
    \subref{fig:spam_CostRatio_mixed_obsv_oracle}, shows the
    corresponding average ratio in $v_e$ over all runs between OBSV
    and the oracle, where for each $\gamma$ the $3^{rd}$~({\wdbc}) /
    $2^{nd}$~({\spam}) extreme values from all runs are denoted by the
    dashed lines. Note a zero value on a logarithmic scale is denoted
    by a cross.  }
  \label{fig:wdbc_spam_random_mixed}
\end{figure*}

The corresponding comparison for the {\bf mixed} sampling strategies is shown in \reffigseptwo{fig:wdbc_CVErrLog_random_mixed}{fig:spam_CVErrLog_random_mixed}.
We again observe the stopping times to increase at a similar rate, and
OBSV to stop earlier on average than the oracle for most values of
$\gamma$
(\reffigseptwo{fig:wdbc_TStop_mixed_obsv_oracle}{fig:spam_TStop_mixed_obsv_oracle}).
Note that the oracle selects the minimum test error at around $180$
labels from {\wdbc} and $1300$ labels from {\spam}, which for both
data sets is only about a half of the number of labels the random
strategy needs.  OBSV tracks these stopping times closely.  Over all,
the fact that in both mixed and random sampling, the stopping times of
OBSV and the oracle are usually well within the extreme value ranges,
indicates a satisfactory performance.

Finally we compare the two sampling strategies directly as shown in
\reffig{fig:wdbc_spam_random_mixed}, using the practical OBSV
algorithm.  As one might expect from the fact that the mixed strategy
converges faster to a low error level, OBSV stops earlier or around
the same time using the mixed strategy than it does for the random
(\reffigseptwo{fig:wdbc_TStop_random_mixed_obsv}{fig:spam_TStop_random_mixed_obsv}).
Those two facts together indicate that OBSV works as intended, since
it stops earlier when convergence is faster.  The results also show
that when using OBSV as a stopping criterion mixed sampling is equal
to or better than random sampling
[\reffigseptwo{fig:wdbc_CostRatio_random_mixed_obsv}{fig:spam_CostRatio_random_mixed_obsv}].
However the differences are mostly not very significant.

\section{Discussion}
\label{sec:discussion}
This paper discussed the interplay between a well-defined cost
function, stopping algorithms and objective evaluation criteria and
their relation to active learning.  Specifically, we have argued that
\begin{inparaenum}[(a)]
\item learning when labels are costly is essentially a stopping problem
\item it is possible to use optimal stopping procedures based on a suitable cost function
\item the goal of active learning algorithms could also be represented
  by this cost function,
\item metrics on this cost function should be used to evaluate
  performance and finally that,
\item the stopping problem cannot be separately considered from
  either the cost function or the evaluation.
\end{inparaenum} 
To our current knowledge, these issues have not yet been sufficiently
addressed.

For this reason, we have proposed a suitable cost function and
presented a practical stopping algorithm which aims to be optimal with
respect to this cost.  Experiments with this algorithm for a specific
prior show that it suffers only small loss compared to the optimal
stopping time and is certainly a step forward from ad-hoc stopping
rules.

On the other hand, while the presented stopping algorithm is an
adequate first step, its combination with active learning is not
perfectly straightforward since the balance between active and uniform
sampling is a hyperparameter which is not obvious how to
set.\footnote{In this paper, the active and uniform sampling rates
  were equal.}  An alternative is to use model-specific stopping
methods.  This could be done if we restrict ourselves to probabilistic
classifiers, as for example in \cite{cohn95active}; in this way we may
be able to simultaneously perform optimal example selection and
stopping.  If such a classifier is not available for the problem at
hand, then judicious use of frequentist techniques such as
bootstrapping~\cite{Efron:1993:Introduction} may provide a
sufficiently good alternative for estimating probabilities.  Such an
approach was advocated by
\cite{icml:roy+mccallum:optimal-active-learning:2001} in order to
optimally select examples; however in our case we could extend this to
optimal stopping.  Briefly, this can be done as follows.
\label{sec:integrated_model}
Let our belief at time $t$ be $\xi_t$, such that for any point $x \in
\CX$, we have a distribution over $\CY$, $\Pr(y \given x, \xi_t)$.  We
may now calculate this over the whole dataset to estimate the realised
generalisation error as the {\em expected error given the empirical
  data distribution and our classifier}
\begin{equation}
  \label{eq:estimated_error}
  \E_D (v_t \given \xi_t) = \frac{1}{|D|} \sum_{i \in D} [1 - \argmax_y \Pr(y_i=y \given x_i, \xi_t)].
\end{equation}
We can now calculate \eqref{eq:estimated_error} for each one of the
different possible labels.  So we calculate the {\em expected error on
  the empirical data distribution if we create a new classifier from
  $\xi_t$ by adding example $i$ } as
\begin{equation}
  \label{eq:estimated_future_error}
  \E_D (v_t \given x_i, \xi_t) = \sum_{y \in \CY}  \Pr(y_i = y \given x_i, \xi_t) \E_D(v_t \given x_i, y_i=y, \xi_t)
\end{equation}
Note that $\Pr(y_i = y \given x_i, \xi_t)$ is just the probability of
example $i$ having label $y$ according to our current belief, $\xi_t$.
Furthermore, $\E_D(v_t \given x_i, y_i=y, \xi_t)$ results from
calculating \eqref{eq:estimated_error} using the classifier resulting
from $\xi_t$ and the added example $i$ with label $y$.  Then
$\E_D(v_t, \xi_t) - \E_D (v_t \given x_i, \xi_t)$ will be the expected
gain from using $i$ to train.  The (subjectively) optimal 1-step
stopping algorithm is as follows: Let $i^* = \argmin_i \E_D (v_t
\given x_i, \xi_t)$.  Stop if $\E_D(v_t \given \xi_t) - \E_D (v_t
\given x_{i^*}, \xi_t) < \gamma$. 

A particular difficulty in the presented framework, and to some extent
also in the field of active learning, is the choice of hyperparameters
for the classifiers themselves.  For Bayesian models it is possible to
select those that maximise the marginal likelihood.\footnote{Other
  approaches require the use of techniques such as cross-validation,
  which creates further complications.}  One could alternatively
maintain a set of models with different hyperparameter choices and
separate convergence estimates.  In that case, training would stop
when there were no models for which the expected gain was larger than
the cost of acquiring another label.  Even this strategy, however, is
problematic in the active learning framework, where each model may
choose to query a different example's label.  Thus, the question of
hyperparameter selection remains open and we hope to address it in
future work.

On another note, we hope that the presented exposition will at the
very least increase awareness of optimal stopping and evaluation
issues in the active learning community, lead to commonly agreed
standards for the evaluation of active learning algorithms, or even
encourage the development of example selection methods incorporating
the notions of optimality suggested in this paper.  Perhaps the most
interesting result for active learning practitioners is the very
narrow advantage of mixed sampling when a realistic algorithm is used
for the stopping times.  While this might only have been an artifact of the
particular combinations of stopping and sampling algorithm and the
datasets used, we believe that it is a matter which should be given
some further attention.

\subsubsection*{Acknowledgements}
We would like to thank Peter Auer for helpful discussions, suggestions
and corrections. This work was supported by the FSP/JRP Cognitive
Vision of the Austrian Science Funds (FWF, Project number S9104-N13).
This work was also supported in part by the IST Programme of the
European Community, under the PASCAL Network of Excellence,
IST-2002-506778.  This publication only reflects the authors' views.

\bibliographystyle{splncs}
\bibliography{./misc}

\end{document}

%% file: preamble.tex
\usepackage{amsmath}
\usepackage{amsfonts}
\usepackage{amssymb}
\usepackage{amsbsy}
\usepackage{theorem}
\usepackage{algorithm}
\usepackage{algorithmic}
\usepackage{mathrsfs}
\usepackage{paralist}
\usepackage{epsfig}
\usepackage{subfigure}
\usepackage{makeidx} 
\usepackage{color}
\usepackage{pstricks}
\usepackage{pst-node}
\usepackage{multido} 
\usepackage{varioref}
\usepackage{wrapfig} %% for wrapping text around figures
\usepackage[numbers]{natbib}

\theoremstyle{plain} 
\theoremstyle{plain} 
\theoremstyle{plain} 
\theoremstyle{plain} 
\theoremstyle{plain} 
\theoremstyle{plain}

\newcommand{\qed}{\nobreak \ifvmode \relax \else
      \ifdim\lastskip<1.5em \hskip-\lastskip
      \hskip1.5em plus0em minus0.5em \fi \nobreak
      \vrule height0.5em width0.5em depth0.25em\fi}

\def \Naturals {{\mathbb{N}}}

\def \CD {{\mathcal{D}}}

\def \CL {{\mathcal{L}}}

\def \CO {{\mathcal{O}}}

\def \CT {{\mathcal{T}}}

\def \CX {{\mathcal{X}}}
\def \CY {{\mathcal{Y}}}

\def\argmax{\mathop{\rm arg\,max}}
\def\argmin{\mathop{\rm arg\,min}}